# A COMPARATIVE ANALYSIS OF REINFORCEMENT LEARNING AND CONVENTIONAL DEEP LEARNING APPROACHES FOR BEARING FAULT DIAGNOSIS

Efe Çakır, Patrick Dumond
Department of Mechanical Engineering, University of Ottawa, Ottawa, Canada
*ecaki090@uottawa.ca

***Abstract*—Bearing faults in rotating machinery can lead to significant operational disruptions and maintenance costs. Modern methods for bearing fault diagnosis rely heavily on vibration analysis and machine learning techniques, which often require extensive labeled data and may not adapt well to dynamic environments. This study explores the feasibility of reinforcement learning (RL), specifically Deep Q-Networks (DQNs) for bearing fault classification tasks in machine condition monitoring to enhance the accuracy and adaptability of bearing fault diagnosis. The results demonstrate that RL models developed in this study can match the performance of traditional supervised learning models under controlled conditions, they excel in adaptability when equipped with optimized reward structures. However, their computational demands highlight areas for further improvement. These findings demonstrate RL's potential to complement traditional methods, paving the way for adaptive diagnostic frameworks.

## I. INTRODUCTION

Reliable bearing performance is essential in industrial settings, as unexpected failures can lead to costly downtime and maintenance interventions. Predicting and preventing bearing failures is, therefore, crucial for maintaining operational efficiency and reducing maintenance costs [1]. This has led to the development and implementation of advanced Prognostics and Health Management (PHM) systems, which aim to monitor, detect, and diagnose faults before they escalate into major failures [2]. By leveraging real-time data and analytical techniques, these systems can provide early warnings of potential issues, allowing for proactive maintenance interventions and minimizing the risk of unexpected bearing failures while maximizing the component useful life [2].

Machine learning (ML), particularly supervised learning (SL), has been extensively applied in PHM for tasks such as fault detection, diagnosis, and prognosis. In SL, algorithms are trained on labeled data, where input-output pairs are known. The goal is to learn a mapping from inputs to outputs, which can be used to make predictions on new, unseen data.

In fault detection, ML techniques such as support vector machines (SVMs) [3] and decision trees [4] have been used to identify patterns in sensor data indicative of anomalies, enabling early detection of potential issues. For fault diagnosis, neural networks and ensemble learning methods are frequently employed to classify different types of faults based on features extracted from vibration data [5].

Reinforcement learning (RL) is a subfield of ML that focuses on training agents to make decisions by interacting with an environment. Unlike SL, where the learning process is guided by labeled data, RL involves learning from trial and error. The agent receives feedback in the form of rewards or penalties based on its actions and uses this feedback to learn a policy that maximizes the cumulative reward.

While RL is not often used for classification tasks, it has potential benefits that can be applied to PHM. Industrial machinery often operates in environments with dynamic variables, fluctuating loads and changing operational conditions. RL's ability to learn and optimize policies in dynamic environments makes it enticing for PHM applications. Furthermore, RL is capable of integrating decision-making into the maintenance process. An RL model can recommend specific maintenance actions based on the current state of the machinery and anticipated future conditions, this improves maintenance scheduling and reduces downtime.

Several researchers have explored RL for fault diagnosis, leveraging its adaptive learning capabilities. Dai et al. [6] developed an RL-based feature selection method using kurtosis to optimize signal-to-noise ratio (SNR) for vibration signals, significantly improving bearing fault detection. Wang et al. [7] introduced the Match-Reinforcement Learning Time-Frequency

Selection (MRLTFS) model, which used RL to autonomously filter and select optimal time-frequency representations, enhancing generalization in small-sample datasets. Mahfooth et al. [8] applied deep reinforcement learning (DRL) for diagnosing short-circuit faults in electric motors. Wang et al. [9] integrated RL with feature engineering to improve robustness against noise and data imbalance, making it effective for diverse fault diagnosis scenarios. While these studies have looked at how to enhance data to improve fault diagnosis, few have tried to use RL for the fault diagnosis task itself. Therefore, this study explores the feasibility of using RL for bearing fault classification tasks in machine condition monitoring.

## II. DATASET AND METHODOLOGY

Although new deep learning approaches emerge regularly, RL is a popular and a powerful tool capable of handling complex tasks that go beyond traditional SLmodels. Its ability to adapt through sequential decision-making makes it particularly versatile. Although RL has not been extensively explored for classification tasks, it holds valuable potential for applications in PHM, especially in fault diagnosis. By integrating RL into this field, the goal of this study is to explore how RL can address fault diagnosis problems and develop an understanding of how it could perform in comparison to the more conventional SL models.

Although RL is better suited for decision-making and interaction-based tasks, its reward-driven approach provides important features that can be leveraged in classification problems like fault diagnosis. This study aims to show how RL can be implemented for fault diagnosis and how future research could refine these models to perform even better.

While the models used here are foundational, they provide insights into how RL can complement traditional SL models in PHM. The inclusion of modified SL models with reward functions and RL environments allows for a thorough comparison of efficiency and accuracy, setting the stage for future improvements that could make RL more viable in fault diagnosis tasks.

### A. Dataset Description

Within-domain performance was evaluated using the UORED dataset [10], which was split into 80% training and 20% testing. Details pertaining to the UORED dataset are provided in Table I. An additional comparison is made by changing the reward matrix and taking advantage of different rewards and penalties for different actions. The reward matrix can be seen in Table VI.

This structure was motivated by the problem of faulty samples getting frequently confused with developing fault samples, and it is achieved by trial and error.

TABLE I. DATASET DESCRIPTION

| Dataset | Date released | Sampling frequency (kHz) | Number of samples | Fault type | Bearing ID |
|---|---|---|---|---|---|
| UORED | 2023 | 42 | 420,000 | Natural | NSK & FAFNIR 6203 |

### B. Model Architectures

Neural network structures are created to be consistent across experiments for ease of comparison, each having layers as described in Table II.

TABLE II. NEURAL NETWORK ARCHITECTURES

| Layer | Type | Input size | Output size | Activation |
|---|---|---|---|---|
| Input Layer | Fully connected | Num_features | 128 | ReLU |
| Hidden Layer | Fully connected | 128 | 128 | ReLU |
| Output Layer | Fully connected | 128 | Num_action | Softmax* |

*For DQN models, the output layer does not use softmax activation, as it directly outputs raw Q-values used for action selection during training. The same is true for models that use the windowing function, as they are handled with CrossEntropy.

To maintain consistency across models and ensure a fair comparative analysis, key hyperparameters, if applicable, were standardized throughout the study and are described in Table III.

TABLE III. HYPERPARAMETERS OF MODELS

| Hyperparameter | Value |
|---|---|
| Batch size | 64 |
| Learning rate | 0.001 |
| Discount factor | 0.99 |
| Epsilon (start) | 1 |
| Decay rate | 0.995 |
| Optimizer | Adam |

These hyperparameters may not be optimal, but they are achieved by trial and error in the pursuit of optimization.

Models in this study, shown in Table IV, include a standard feedforward ANN that represents a standard SL model, a modified ANN with reward-based training that incorporates a reward-based loss function, and two types of Deep Q-Networks (DQNs) that represent RL. The difference between the two DQNs is the inclusion of replay buffer handling. Also, an alternative DQN includes timestep-based training rather than epoch-based updates.

TABLE IV. MODELS USED IN THIS STUDY

| Data upload method / Type of model | Within Domain Analysis | |
|---|---|---|
| | *Windows of raw data* | *Statistical feature vectors* |
| ANN | Model 1 | Model 5 |
| ANN modified with RL environment and reward function | Model 2 | Model 6 |
| DQN | Model 3 and 4 | Model 7 and 8 |

### C. Feature Selections

Given the nature of the vibration data, two primary preprocessing approaches were employed: windowing and statistical feature extraction. Each preprocessing method is designed to represent the data differently, providing inputs to the models for both within-domain and cross-domain evaluations.

*1) Windowing Data Augmentation*

In the windowing data augmentation method, the raw vibration data is divided into fixed-length windows of 1000 data points. 50% overlap between consecutive windows [11] ensures that no significant information is lost. The process can be seen in Fig. 1. This approach retains the time-domain structure of the vibration signal, enabling models to learn from the raw temporal patterns present in the data.

The windowed data is used as an input to both the RL and SL models used in this study, with the windowing technique allowing the models to focus on localized time-domain variations.

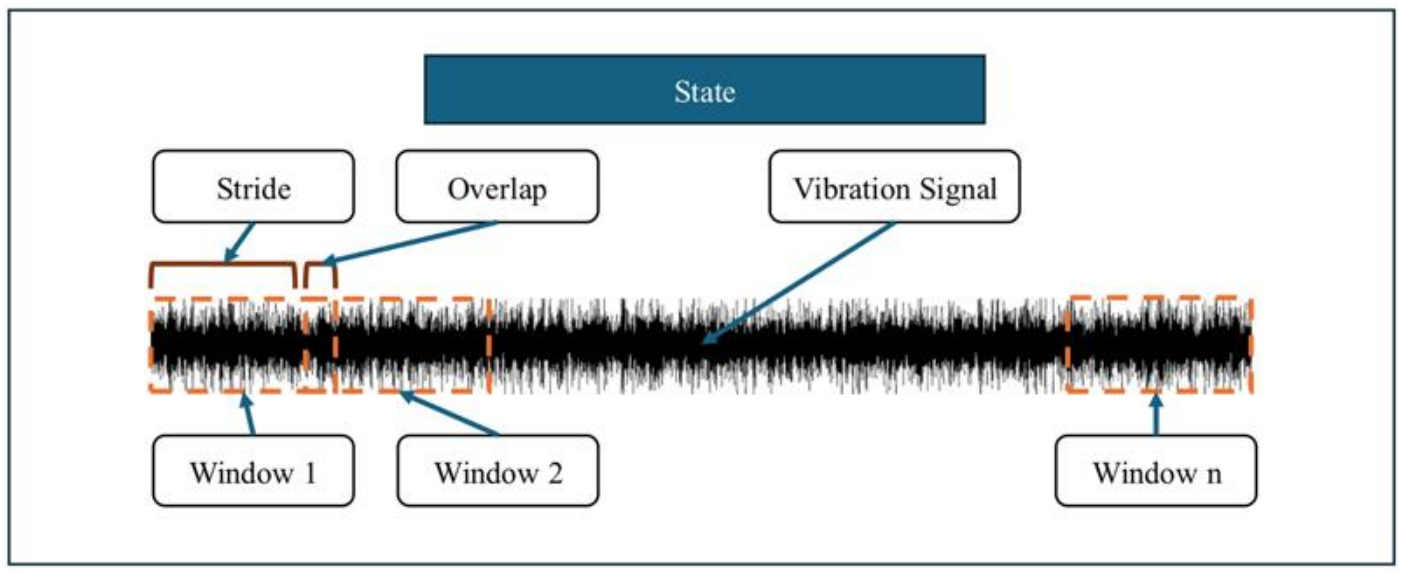

Figure 1. Windowing data augmentation diagram

*2) Statistical Feature Extraction*

In this method, a set of statistical features is extracted from the raw vibration data to summarize the key characteristics of each data file or parts of the data files. These features include mean, standard deviation, peak-to-peak, RMS, skewness, kurtosis, crest factor, shape factor, impulse factor, margin factor, and peak factor.

For each segment of the files, these statistical features are extracted and stored in a “statistical feature vector”, reducing the raw signal data into a more compact form. This process is depicted in Fig. 2. These feature vectors are used as inputs for both RL and SL models, allowing for a more abstract representation of the data.

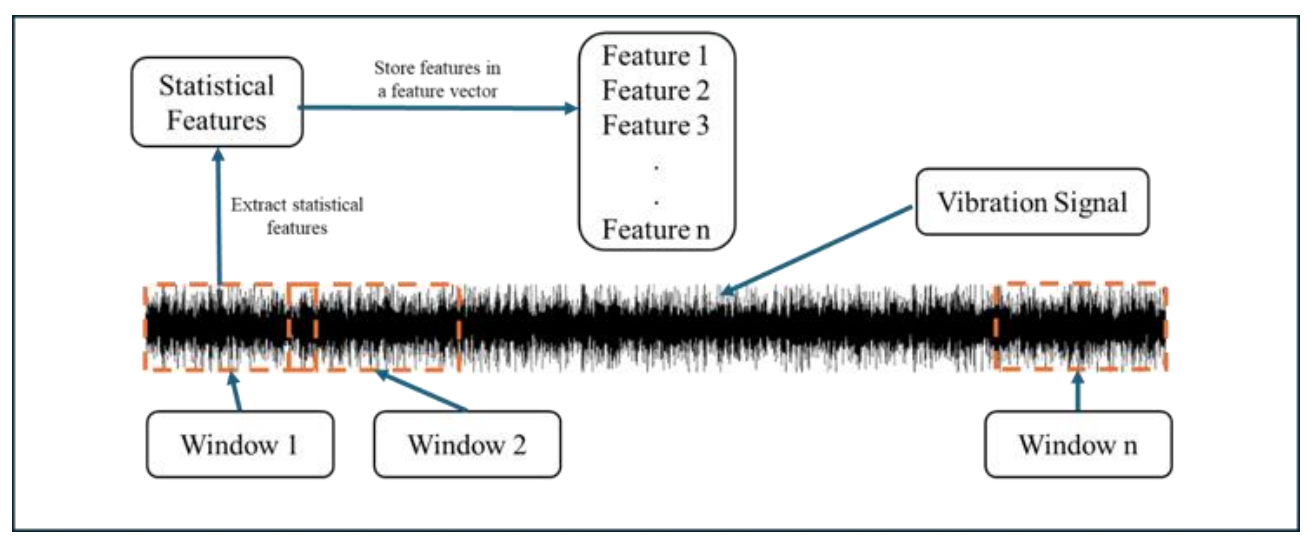

Figure 2. Statistical feature extraction diagram

## III. Results and Discussion

### A. UORED Dataset Results for Windowing Data Augmentation

Model architectures for models 1, 2, and 3 are kept consistent in terms of how data is loaded and trained. Due to the poor performance of model 3, an alternative model (model 4) was created to see if the poor performance of the DQN was due to its architecture, which resembled that of the SL model.

Unlike the other models that trained with epochs, Model 4 was trained using timesteps, which defines a discrete step each time the model observes and interacts with the environment. The DQN in model 4 benefits from this structure by utilizing a replay buffer more effectively, storing past experiences and allowing it to learn from them across timesteps. This setup, with a total of 160,000 timesteps, enables the DQN to sample from diverse experiences, improving the model's performance by balancing exploration and exploitation in a way that differs from epoch-based training used in SL models. The performances of models for this analysis can be found in Table V.

TABLE V. Performance of Models Trained with Windows of Raw Data

| Model | Accuracy (%) | Precision | Recall | F-1 Score | Relative training time |
|---|---|---|---|---|---|
| Model 1: ANN | 67.45 | 0.67 | 0.67 | 0.67 | 6m 17.3s |
| Model 2: Modified ANN | 66.24 | 0.64 | 0.66 | 0.65 | 6m 28.7s |
| Model 3: DQN | 33.06 | 0.29 | 0.33 | 0.20 | 642m 24.9s |
| Model 4: Alternative DQN | 62.57 | 0.62 | 0.64 | 0.63 | 25m 42.1s |

Table V shows that models 1, 2, and 4 performed similarly, with model 1 yielding slightly better results. Model 3 performed very poorly even with its immense training time, which is close to 11 hours. Model 4, which has a different architecture, was able to create a decent policy and performed as well as the SL models.

This modification of the DQN arose from the need to store experiences more efficiently. Models using windowing data augmentation have many large windows of raw data, which challenged the DQN's structure. While SL models have direct access to the entire dataset they are trained with, RL models learn gradually from experiences stored in the replay buffer. This fundamental difference requires adjustments to the data loading method and an experience replay strategy to optimize the DQN for improved performance. Confusion matrices for the models can be found in Fig. 3.

This modification was not necessary for models trained with statistical feature vectors, as these vectors were fewer in number and inherently more compact compared to windows containing 1,000 raw data points. Statistical feature vectors provided a condensed representation of the data, making them manageable for standard processing without requiring adjustments to the replay buffer or data handling methods.

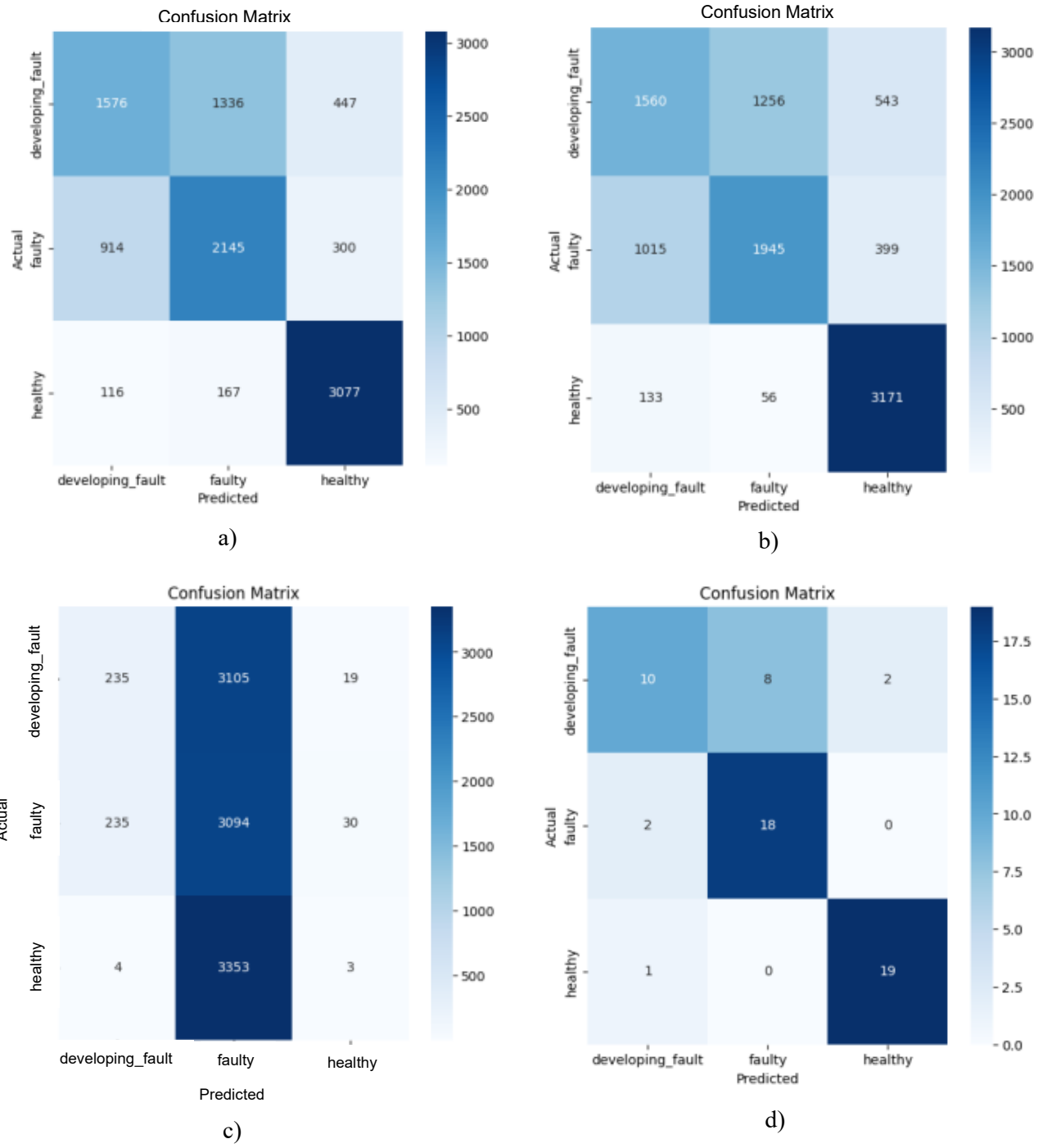

Figure 3. Confusion matrices of a) standard ANN, b) modified ANN, c) DQN, d) alternate DQN for windowing data augmentation

## *B. UORED Dataset Results with Statistical Feature Vectors*

Structures of all models in Table VII are kept consistent. An additional test is done to see if there is a performance boost that can be achieved by implementing different rewards and penalties for individual decisions. The structure of the tested reward mechanism can be found in Table VI. This structure was motivated by the problem of faulty samples getting frequently confused with developing fault samples, and it is achieved by trial and error.

TABLE VI. IMPLEMENTED REWARD MATRIX

| | | Predicted | | |
|---|---|---|---|---|
| | | *Developing fault* | *Faulty* | *Healthy* |
| **Actual** | *Developing fault* | +1 | -1.2 | -1 |
| | *Faulty* | -1.2 | +1 | -0.5 |
| | *Healthy* | -1 | -0.5 | +1 |

This additional experiment was conducted because the first 3 models confused faulty and developing fault samples, as can be observed in Fig. 5. The performances of models that are trained via statistical feature vectors can be found in Table VII.

Results of these trials favor the DQN. We can see a slight increase of about 4% on all evaluation metrics for the DQN compared to the ANN. This increase is even more significant with the updated reward mechanism, with about 8% difference on all metrics.

The alternative reward matrix that is tailored to the performance of the DQN with the UORED dataset outperformed the initial DQN as it accumulated more rewards overall, as can be seen in Fig. 4. This approach also improved the problem of developing fault samples getting confused with faulty samples, as can be seen in the confusion matrices of Fig. 5.

TABLE VII. PERFORMANCE OF MODELS TRAINED WITH STATISTICAL FEATURE VECTORS

| Model | Accuracy (%) | Precision | Recall | F-1 Score | Relative training time |
|---|---|---|---|---|---|
| Model 5: ANN | 70.00 | 0.69 | 0.70 | 0.70 | 2.5s |
| Model 6: Modified ANN | 71.67 | 0.72 | 0.72 | 0.72 | 2m 56.2s |
| Model 7: DQN | 73.33 | 0.74 | 0.73 | 0.74 | 5m 2.0s |
| Model 8: DQN with different rewards | 78.33 | 0.79 | 0.78 | 0.77 | 5m 38.0s |

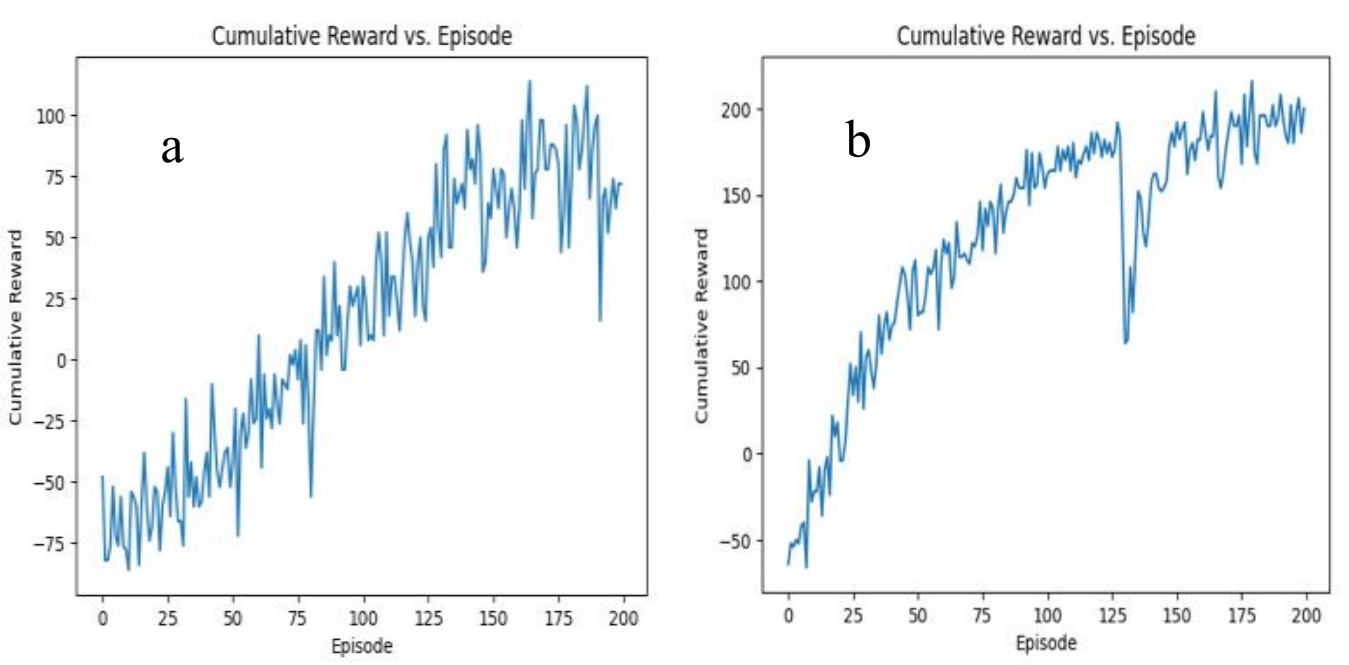

Figure 4. Cumulative reward vs. episode for a) DQN with preset rewards, b) DQN with alternative rewards

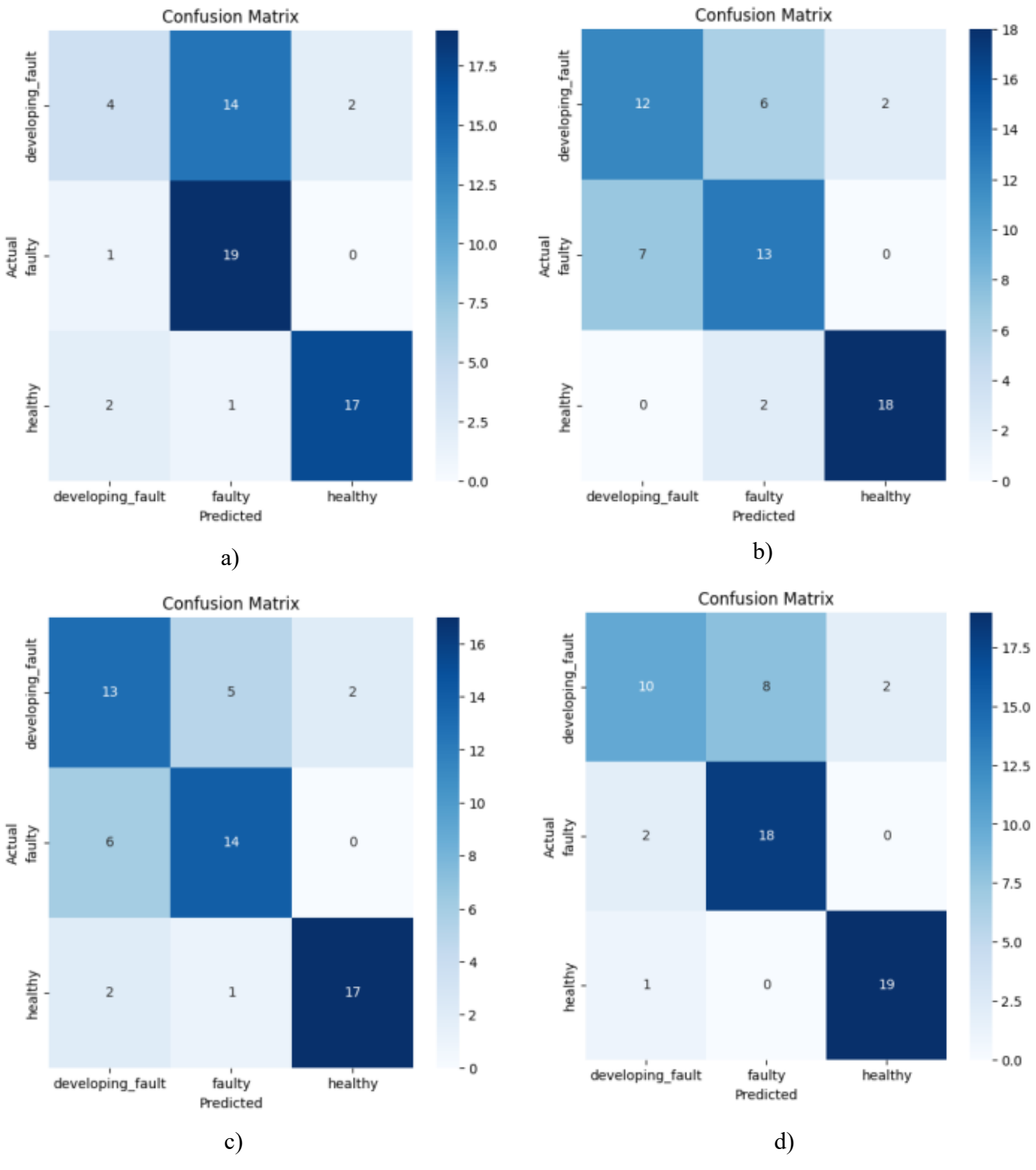

Figure 5. Confusion matrices of a) standard ANN, b) modified ANN, c) DQN, d) DQN with different rewards for statistical feature vectors

A similar approach can be taken in SL models by adjusting class weights – which makes the model biased towards classifying certain classes. However, the reward functions in RL models are more versatile as rewards or penalties can be applied to any action the agent makes.

Overall results show that RL models face both distinct challenges and opportunities compared to SL models in bearing fault diagnosis. When trained with windowed data, RL models struggled due to the complexity and high dimensionality of the input, leading to longer training times and lower performance unless modifications are introduced. Conversely, RL models performed significantly better with statistical feature vectors, as the reduced dimensionality enabled efficient policy learning and improved accuracy, particularly when a tailored reward matrix was implemented.

## IV. Conclusions

This study explored the feasibility of implementing RL for bearing fault diagnosis by comparing its performance with traditional SL approaches. By integrating RL-based models into the diagnostic process, this study demonstrates the potential of RL for fault detection tasks while also highlighting the challenges that must be overcome to make RL a viable alternative to SL in PHM. The results show that while SL methods remain sufficiently effective for bearing fault classification, RL approaches offer promising adaptability and performance. RL could be used as a powerful tool that complements SL models at tackling more complex tasks.

## References

[1] R. Rejith, D. Kesavan, P. Chakravarthy, and S. V. S. Narayana Murty, "Bearings for aerospace applications," *Tribology International*, vol. 181, p. 108312, Mar. 2023, doi: 10.1016/j.triboint.2023.108312.

[2] E. Zio, "Prognostics and Health Management (PHM): Where are we and where do we (need to) go in theory and practice," *Reliability Engineering & System Safety*, vol. 218, p. 108119, Feb. 2022, doi: 10.1016/j.ress.2021.108119.

[3] Y. Yang, D. Yu, and J. Cheng, "A fault diagnosis approach for roller bearing based on IMF envelope spectrum and SVM," *Measurement*, vol. 40, no. 9, pp. 943–950, Nov. 2007, doi: 10.1016/j.measurement.2006.10.010.

[4] A. Muniyappa, V. Sugumaran, and H. Kumar, "Exploiting sound signals for fault diagnosis of bearings using decision tree," *Measurement*, vol. 46, pp. 1250–1256, Apr. 2013, doi: 10.1016/j.measurement.2012.11.011.

[5] Y. Li, Y. Zhang, R. Wang, and J. Fu, "A Reinforcement Ensemble Learning Method for Rolling Bearing Fault Diagnosis under Variable Work Conditions," *Sensors*, vol. 24, no. 11, p. 3323, May 2024, doi: 10.3390/s24113323.

[6] W. Dai, Z. Mo, C. Luo, J. Jiang, and Q. Miao, "Bearing Fault Diagnosis Based On Reinforcement Learning And Kurtosis," in *2019 Prognostics and System Health Management Conference (PHM-Qingdao)*, Qingdao, China: IEEE, Oct. 2019, pp. 1–5. doi: 10.1109/PHM-Qingdao46334.2019.8942977.

[7] J. Wang *et al.*, "Match-reinforcement learning with time frequency selection for bearing fault diagnosis," *Meas. Sci. Technol.*, vol. 34, no. 12, p. 125005, Dec. 2023, doi: 10.1088/1361-6501/ace644.

[8] M. Mahfooth and O. Mahmood, "Detection and Diagnosis of Inter-Turn Short Circuit Faults of PMSM for Electric Vehicles Based on Deep Reinforcement Learning," *(AREJ)*, vol. 28, no. 2, pp. 75–85, Sep. 2023, doi: 10.33899/rengj.2023.138195.1229.

[9] R. Wang, H. Jiang, K. Zhu, Y. Wang, and C. Liu, "A deep feature enhanced reinforcement learning method for rolling bearing fault diagnosis," *Advanced Engineering Informatics*, vol. 54, p. 101750, Oct. 2022, doi: 10.1016/j.aei.2022.101750.

[10] M. Sehri and P. Dumond, "University of Ottawa Rolling-element Dataset – Vibration and Acoustic Faults under Constant Load and Speed conditions – Spectrograms (UORED-VAFCLS-P1) - Mendeley Data." Apr. 04, 2023. doi: 10.17632/65d7pmfzvx.

[11] M. Sehri and P. Dumond, *Optimizing Rolling Element Bearing Data Collection and Algorithm Hyperparameters for Machine Learning*. 2024. doi: 10.2139/ssrn.4949181.